\documentclass{article}
\usepackage[T1]{fontenc}

\usepackage{graphicx}

\usepackage{url}

\usepackage{authblk}

\usepackage{subfig}
\usepackage{amsmath}

\usepackage{hyperref}

\usepackage{tikz}
\usetikzlibrary{shapes.geometric,shapes.arrows,calc,positioning,arrows.meta,bending}
\usepackage{xcolor}

\usepackage{listings}

\definecolor{blue}{HTML}{4472c4}
\definecolor{red}{HTML}{c00000}
\definecolor{green}{HTML}{70ad47}

\tikzset{mathmoddbCircle/.style={draw=black, thick, fill=blue, circle, minimum size=3.25cm, text=white, align=center}}

\tikzset{mathalgodbCircle/.style={mathmoddbCircle, fill=red}}
\tikzset{mathdbCircle/.style={mathmoddbCircle, fill=green}}
\tikzset{labelBox/.style={fill=lightgray, draw=none, rectangle, rounded corners, inner sep=2pt, align=center, 
										minimum width = 3.25cm, minimum height = 0.6cm}
	}

\tikzset{mathdbArrow/.style n args={1}%
	{to path={let \p1 = ($(\tikztotarget)-(\tikztostart)$),
			\n1 = {int(mod(scalar(atan2(\y1,\x1))+360, 360))},
			\n2 = {veclen(\x1,\y1)}
			in \pgfextra{\typeout{\n1,\n2,\x1,\y1}}
			(\tikztotarget)
			node[single arrow, minimum height=\n2-\pgflinewidth, inner sep=3pt, single arrow head extend=1ex, rotate=\n1, anchor=tip, fill=white, draw=blue, thick]{#1}
}}}

\tikzset{mathdbArrow2/.style n args={1}%
	{to path={let \p1 = ($(\tikztotarget)-(\tikztostart)$),
			\n1 = {int(mod(scalar(atan2(\y1,\x1))+360, 360))},
			\n2 = {veclen(\x1,\y1)}
			in \pgfextra{\typeout{\n1,\n2,\x1,\y1}}
			(\tikztotarget)
			node[single arrow, minimum height=\n2-\pgflinewidth, inner sep=3pt, single arrow head extend=1ex, text width=3cm, align=center, rotate=\n1, anchor=tip, fill=white, draw=blue, thick]{#1}
}}}

\tikzset{mathdbArrowEquiv/.style n args={1}%
	{to path={let \p1 = ($(\tikztotarget)-(\tikztostart)$),
			\n1 = {int(mod(scalar(atan2(\y1,\x1))+360, 360))},
			\n2 = {veclen(\x1,\y1)} in \pgfextra{\typeout{\n1,\n2,\x1,\y1}} (\tikztotarget)
			node[double arrow, double arrow head extend=1ex, minimum height=\n2-\pgflinewidth, inner sep=3pt, rotate=\n1, anchor=east, fill=white, draw=blue, thick]{#1}
}}}

\usepackage{fancyhdr}

\fancypagestyle{plain}{%
   \fancyhf{} 
\fancyfoot[C]{\footnotesize Preprint submitted to the \textit{18th International Conference on Metadata and Semantics Research} 2024 and published as a full, revised article in \textit{Sfakakis, M., Garoufallou, E., Damigos, M., Salaba, A., Papatheodorou, C. (eds) Metadata and Semantic Research. MTSR 2024. Communications in Computer and Information Science, vol 2331. Springer, Cham.} \url{https://doi.org/10.1007/978-3-031-81974-2_8}}

}

\begin{document}
\title{Towards a Knowledge Graph for Models and Algorithms in Applied Mathematics}

\author[1]{Björn Schembera\thanks{Corresponding author: \textit{bjoern.schembera@ians.uni-stuttgart.de}}}
\author[2]{Frank Wübbeling}
\author[2]{Hendrik Kleikamp}
\author[3]{Burkhard Schmidt}
\author[3]{Aurela Shehu}
\author[4]{Marco Reidelbach}
\author[5]{Christine Biedinger}
\author[5]{Jochen Fiedler}

\author[3]{Thomas Koprucki}
\author[6]{Dorothea Iglezakis}
\author[1,7]{Dominik Göddeke}

\affil[1]{\small{Institute of Applied Analysis and Numerical Simulation, University of Stuttgart}}
\affil[2]{\small{Institute of Applied Mathematics: Analysis and Numerics, University of Münster}}
\affil[3]{\small{Weierstrass Institute for Applied Analysis and Stochastics, Berlin}}

\affil[4]{\small{Mathematics of Complex Systems, Zuse Institute Berlin}}
\affil[5]{\small{Fraunhofer Institute for Industrial Mathematics, Kaiserslautern}}
\affil[6]{\small{University Library, University of Stuttgart}}
\affil[7]{\small{Stuttgart Center for Simulation Science (SC SimTech), University of Stuttgart}}

\maketitle              
\begin{abstract}
Mathematical models and algorithms are an essential part of mathematical research data, as they are epistemically grounding numerical data. In order to represent models and algorithms as well as their relationship semantically to make this research data FAIR, two previously distinct ontologies were merged and extended, becoming a living knowledge graph. The link between the two ontologies is established by introducing computational tasks, as they occur in modeling, corresponding to algorithmic tasks.
Moreover, controlled vocabularies are incorporated and a new class, distinguishing base quantities from specific use case quantities, was introduced. Also, both models and algorithms can now be enriched with metadata. Subject-specific metadata is particularly relevant here, such as the symmetry of a matrix or the linearity of a mathematical model. This is the only way to express specific workflows with concrete models and algorithms, as the feasible solution algorithm can only be determined if the mathematical properties of a model are known. We demonstrate this using two examples from different application areas of applied mathematics. In addition, we have already integrated over 250 research assets from applied mathematics into our knowledge graph.

\end{abstract}
\section{Introduction}

Data and knowledge driven approaches constitute the fourth pillar of science. Computer simulations generate vast amounts of data, big measurement data are recorded in physics, and statistical data are collected in social science cohort studies, to name some examples. In general, for scientific reasoning, the processing and generation of data is becoming increasingly important. Moreover, sharing and citing research data is more and more acknowledged as a key part of the scientific process~\cite{Conrad2024}.
To enable this and to ensure good scientific practice, research data must be documented and stored in accordance with the FAIR principles~\cite{Wilkinson2016} as a means to avoid dark data~\cite{Schembera2020_DD}. In particular, all relevant information that has led to a scientific finding must be documented in order to be able to replicate and reproduce the result at a later date or by third parties~\cite{Baker2016,Riedel2022}.

Specifically in branches of science using mathematical methods, research data take many forms. 
Mathematics is typically still associated with classical mathematical artifacts, namely documents with mathematical proofs and formulae. However, this definition falls short, as applied mathematics produce huge amounts of data based on numerical methods. These particular fields heavily rely on mathematical models and algorithms for the generation of numerical or symbolic data~\cite{Boege2022,MaRDI2022}. A comprehensive overview of types of mathematical research data can be found in~\cite{Conrad2024}.

For a complete epistemic understanding, all models and solution algorithms must be documented. In order to tackle similar problems or to improve efficiency, alternate, competing or complementing models and solution schemes can be identified from this knowledge, which goes beyond reproducibility.
The conceptual foundations for semantic knowledge representation for these research assets were laid in~\cite{schembera2024ontologies,Schembera2023_CoRDI}, 
where an ontology for mathematical models and algorithms was proposed.  
Building on top of this conceptual work in~\cite{schembera2024ontologies}, we present a matured version of a joint ontology of models and algorithms stemming from applied mathematics, and a knowledge graph (KG) with to date more than 2000 elements, ready for production service.
To this end, the paper is structured as follows: Sec.~\ref{sec:RelatedWork} provides an overview of relevant related work. Sec.~\ref{sec:MergingModAlgo} first discusses the preliminary work on which our approach is based on, then details our extensions and the merging of the ontologies. Sec.~\ref{sec:KnowledgeGraph} afterwards describes the transition from the joint ontology to a KG with data using examples. Sec.~\ref{sec:ConclusionOutlook} concludes the paper with a summary and an outlook.

\section{Related Work}
\label{sec:RelatedWork}

To date, the use and development of ontologies in mathematics is not yet wide\-spread, as the realization that all research data are relevant and should be made FAIR has only recently become more commonly accepted.

Preliminary work in the area of mathematical models exists, e.g., introducing taxonomic classifications of mathematical artifacts~\cite{Elizarov2017}, which were eventually merged with succeeding educational approaches~\cite{Falileeva2020,Kirillovich2021}. Other didactic uses of ontologies for mathematical learning  can be found in~\cite{Zang2022,Zwaneveld2000}. However, since these ontologies are primarily educational and taxonomic, they do not include many mathematical objects relevant to our work.

In addition, there are many subject-specific ontologies for mathematical models from a certain domain, e.g., plasma physics~\cite{Snytnikov2020}, biology~\cite{Chelliah2013,Inizan2021}, Neural Networks~\cite{Nguyen2020}, in a mechanistic context~\cite{Suresh2010,Suresh2008} or for models built on interval data~\cite{Dyvak2022}. In contrast, our approach is broader and more flexible and aims to create a general ontology for mathematical models and (numerical) algorithms. Our ontology is designed to be modular to possibly connect with other ontologies and knowledge frameworks, such as those for software and hardware, in the context of Linked Open Data
\footnote{\url{https://www.w3.org/DesignIssues/LinkedData.html}}.

Based on the assumption that mathematical papers hold research data, Open Mathematical Documents
~\cite{Kohlhase:OMDoc} introduce a semantic markup format defining an ontology for mathematical documents. 
A more general approach to representing these semantics is the use of \textit{Model Pathway Diagrams} (MPDs)~\cite{Koprucki2018}. 

MPDs can be viewed as preliminary work leading to our development of an ontology for mathematical models and algorithms.

The Algorithms Metadata Vocabulary
~\cite{Dutta2022} is designed from a computer science perspective 
presenting detailed knowledge about algorithms, including characteristics such as loop constructs or data structures. The MEX Algorithm Ontology
~\cite{Esteves2015} focuses on representing the algorithmic information present in a basic machine learning experiment. While these details are crucial for implementers, computer and ML scientists, our focus is solely on mathematical algorithms.

The work presented is being driven by the \textit{Mathematical Research Data Initiative} (MaRDI)~\cite{MaRDI2022}, a consortium within the \textit{National German Research Data Infrastructure} (NFDI)~\cite{Hartl2021}. The goal of the NFDI is a linked data infrastructure and semantic technology built by discipline specific consortia. 
The NFDI Core Ontology~\cite{Sack2023} describes and connects research outputs like datasets, software as well as events with agents and their roles and scientific domains. 
The Open Research Knowledge Graph (ORKG)~\cite{Auer2018} describes scientific papers of different domains in a semantic way, making it easier to find and compare.  
As mathematical models and algorithms are intensively used in engineering, the ontology metadata4ing~\cite{Arndt2022_m4i}, allows to connect research outputs with agents, methods, tools and the object of research via provenance tracking.  
wn that the MathModDB works beyond its original domain, when used to document an, albeit simple, algebraic \textit{Object Comparison Model}. Further investigations are necessary to conclude whether or not MathModDB is able to handle more complex algebraic models and models from other domains.

\section{Merging MathModDB and MathAlgoDB}

\label{sec:MergingModAlgo}

In this section, we review the previously introduced ontologies for mathematical models and algorithms laid out in~\cite{schembera2024ontologies,Schembera2023_CoRDI} and present an extension and conceptual redesign. The ontology described in this section acts as a data model for the KG described in Sec.~\ref{sec:KnowledgeGraph}. Please note that the KG AlgoData in~\cite{schembera2024ontologies} has since been renamed to MathAlgoDB.

\subsection{Previous Ontology Structures and their Shortcomings}
\label{subsec:PrevWork}
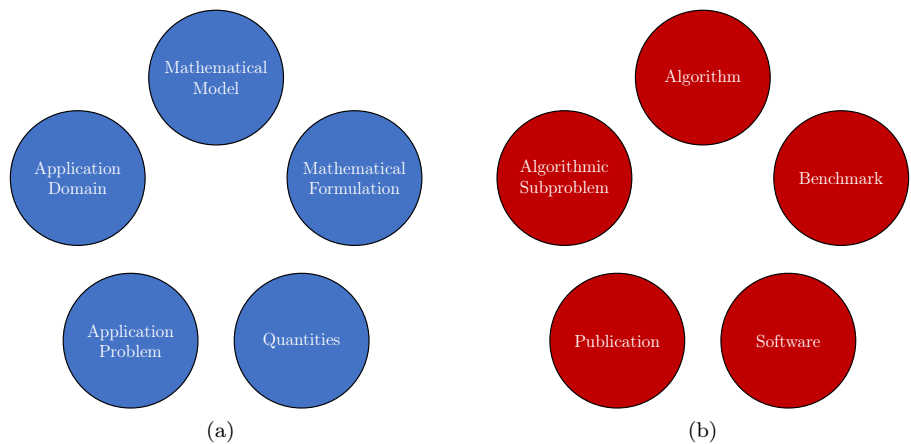
\begin{figure}[!ht]
    \centering
   \subfloat[]{
        \resizebox{0.45\textwidth}{!}{%
            \begin{tikzpicture}
                \node[minimum size=7cm, regular polygon, regular polygon sides=5] (a) {};
        		\node[mathmoddbCircle] (matModel) at (a.corner 1) {\large Mathematical\\[2pt]\large Model};
        		\node[mathmoddbCircle] (matForm) at (a.corner 5) {\large Mathematical\\[2pt]\large Formulation};
        		\node[mathmoddbCircle] (quan) at (a.corner 4) {\large Quantities};
        		\node[mathmoddbCircle] (applProb) at (a.corner 3) {\large Application\\[2pt]\large Problem};
        		\node[mathmoddbCircle] (applDomain) at (a.corner 2) {\large Application\\[2pt]\large Domain};
        	\end{tikzpicture}%
         }
        \label{fig:MathModDBSchematic}
    }
    \hfill
    \subfloat[]{
        \resizebox{0.45\textwidth}{!}{%
            \begin{tikzpicture}
                \node[minimum size=7cm, regular polygon, regular polygon sides=5] (a) {};
        		\node[mathalgodbCircle] (matModel) at (a.corner 1) {\large Algorithm};
        		\node[mathalgodbCircle] (matForm) at (a.corner 5) {\large Benchmark};
        		\node[mathalgodbCircle] (quan) at (a.corner 4) {\large Software};
        		\node[mathalgodbCircle] (applProb) at (a.corner 3) {\large Publication};
        		\node[mathalgodbCircle] (applDomain) at (a.corner 2) {\large Algorithmic\\[2pt]\large Subproblem};
        	\end{tikzpicture}%
         }
        \label{fig:MathAlgoDBSchematic}
    }
    \caption{Essential classes of the  related work ontologies for mathematical models MathModDB (a) and numerical algorithms MathAlgoDB (b) from~\cite{schembera2024ontologies}.}
\end{figure}

In~\cite{schembera2024ontologies}, mathematical models have been identified as one of the central categories of mathematical research data and the \textit{Mathematical Models Ontology} (MathModDB) has been developed, with its essential classes shown in Fig.~\ref{fig:MathModDBSchematic}.
The same holds true for numerical algorithms, leading to the \textit{MathAlgoDB} ontology with its essential classes shown in Fig.~\ref{fig:MathAlgoDBSchematic}.

These class names showed some inconsistencies with other common standard nomenclature, such as Wikidata\footnote{\url{https://www.wikidata.org/wiki/Q2465832}} and the ORKG\footnote{\url{https://orkg.org/fields}}. 
Hence, minor modifications of the original design were made by renaming the classes \textit{Application Domain} to \textit{Research Field} and \textit{Application Problem} to \textit{Research Problem}.

Major adjustments became necessary as a result of the following considerations. Solving a scientific problem derived from an application problem via modeling (represented in MathModDB) requires addressing algorithmic subproblems (represented in MathAlgoDB). In~\cite{schembera2024ontologies}, this connection was made through the relation \textit{uses algorithmic problem} and its inverse \textit{used by model problem},

representing the process where mathematical modeling is followed by solving model equations with a numerical algorithm via the class of \textit{Algorithmic Problem}s. In~\cite{schembera2024ontologies}, it was already mentioned that distinguishing between a mathematical problem and an algorithmic subproblem can be challenging. Sometimes, specific \textit{Mathematical Model}s in MathModDB had no corresponding \textit{Algorithmic Problem}s in MathAlgoDB. In such cases, the mathematical problem needed to be added to MathAlgoDB as an \textit{Algorithmic Problem}. In this respect, there are strong canonical connections between the two ontologies, which were further specified in the following active development. 

\subsection{Computational Tasks as the missing Link}
\label{subsec:CompTask}
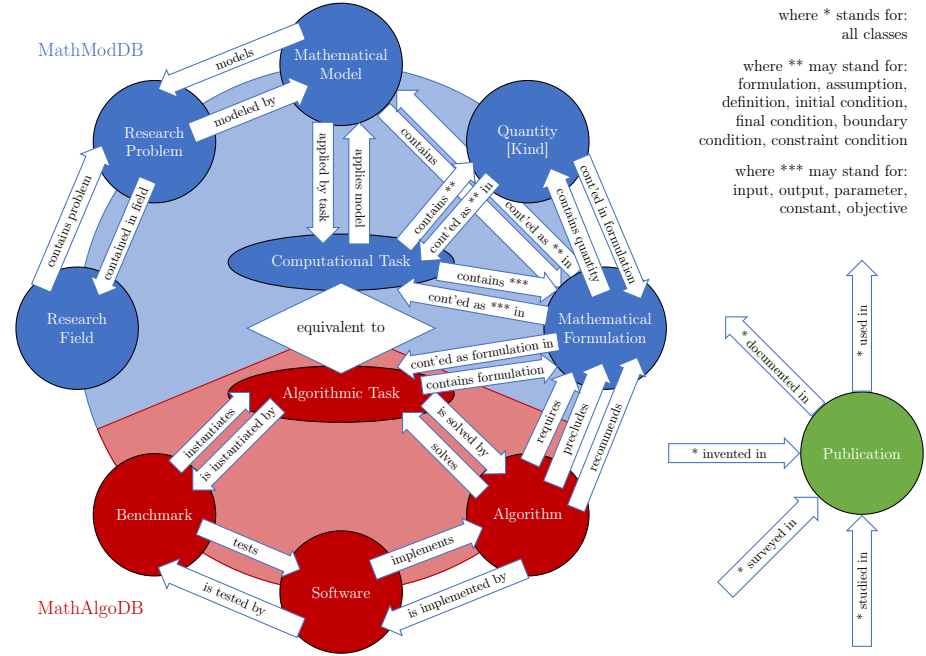
\begin{figure}[!ht]
    \centering

    \resizebox{\textwidth}{!}{
        \begin{tikzpicture}
    		\node[minimum size=14cm, regular polygon, regular polygon sides=8, rotate=22.5] (a) {};
    		\draw[blue, thick, fill=blue!50] (a.center) circle (7cm);
    		\draw (79.5:5.85) edge[mathdbArrow={contains\hphantom{some nice phantom text}}] (10.5:5.95) {};
    		\draw (13.5:6.5) edge[mathdbArrow={\rotatebox{180}{cont'ed as ** in}\hphantom{some nice phantom text}}] (77:6.5) {};
    		\draw[red, thick, fill=red!50] (a.center) --  (337.5:7) arc(337.5:202.5:7) -- cycle;
    		\node[mathmoddbCircle] (matModel) at (a.corner 1) {\large Mathematical\\[2pt]\large Model};
    		\node[mathmoddbCircle] (quan) at (a.corner 8) {\large Quantity\\[2pt]\large [Kind]};
    		\node[mathmoddbCircle] (matForm) at (a.corner 7) {\large Mathematical\\[2pt]\large Formulation};
    		\node[mathalgodbCircle] (soft) at (a.corner 5) {\large Software};
    		\node[mathalgodbCircle] (algo) at (a.corner 6) {\large Algorithm};
    		\node[mathalgodbCircle] (bench) at (a.corner 4) {\large Benchmark};
    		\node[mathmoddbCircle] (reseField) at (a.corner 3) {\large Research\\[2pt]\large Field};
    		\node[mathmoddbCircle] (reseProb) at (a.corner 2) {\large Research\\[2pt]\large Problem};
    		\node[mathmoddbCircle, minimum width=6cm, ellipse, minimum height=1.5cm] (compTask) at ($(a.center) + (0, 1.75cm)$) {\large Computational Task};
    		\node[mathalgodbCircle, minimum width=6cm, ellipse, minimum height=1.5cm] (algoTask) at ($(a.center) - (0, 1.75cm)$) {\large Algorithmic Task};
    		\node[diamond, draw=blue, thick, minimum width=5cm, inner sep=-2pt, fill=white] at (a.center) {\large equivalent to};
    
    		\draw (127.5:6.5) edge[mathdbArrow={modeled by}] (97.5:6.5) {};
    		\draw (97.5:8) edge[mathdbArrow={\rotatebox{180}{models}}] (127.5:8) {};
    		\draw (172.5:8) edge[mathdbArrow={contains problem}] (142.5:8) {};
    		\draw (142.5:6.5) edge[mathdbArrow={\rotatebox{180}{contained in field}}] (172.5:6.5) {};
    		\draw (35.5:7.8) edge[mathdbArrow={cont'ed in formulation}] (5.5:8) {};
    		\draw (7.5:7) edge[mathdbArrow={\rotatebox{180}{contains quantity}}] (37.5:7) {};
    		\draw (307.5:8) edge[mathdbArrow={\rotatebox{180}{is implemented by}}] (277.5:8) {};
    		\draw (277.5:6.5) edge[mathdbArrow={implements}] (305.5:6.5) {};
    		\draw (262.5:8) edge[mathdbArrow={\rotatebox{180}{is tested by}}] (232.5:8) {};
    		\draw (234.5:6.5) edge[mathdbArrow={tests}] (260.5:6.5) {};
    		\draw (95:5.5) edge[mathdbArrow={applied by task}] ($(95:5.5) - (0, 3.25cm)$) {};
    		\draw ($(85:5.5) - (0, 3.25cm)$) edge[mathdbArrow={\rotatebox{180}{applies model}}] (85:5.5) {};
    		\draw (52.5:2.75) edge[mathdbArrow={contains **}] (51.5:5.65) {};
    		\draw (45:5.65) edge[mathdbArrow={\rotatebox{180}{cont'ed as ** in}}] (40:2.75) {};
    		\draw (229.5:2.95) edge[mathdbArrow={\rotatebox{180}{is instantiated by}}] (227.5:5.85) {};
    		\draw (220:5.75) edge[mathdbArrow={instantiates}] (215:2.95) {};
    		\draw (320.5:2.95) edge[mathdbArrow={is solved by}] (318.5:5.85) {};
    		\draw (311:5.75) edge[mathdbArrow={\rotatebox{180}{solves}}] (306:2.75) {};
    		\draw (30:2.95) edge[mathdbArrow={contains ***}] (10:5.85) {};
    		\draw (4:5.5) edge[mathdbArrow={\rotatebox{180}{cont'ed as *** in}}] (35:1.8) {};
    		\draw (330:2.95) edge[mathdbArrow={contains formulation}] (350:5.85) {};
    		\draw (-4:5.5) edge[mathdbArrow={\rotatebox{180}{cont'ed as formulation in}}] (-35:1.8) {};
    
    		\draw (-37.5:7.9) edge[mathdbArrow={recommends}] (-5.5:7.9) {};
    		\draw (-36.5:7) edge[mathdbArrow={precludes}] (-7.5:7) {};
    		\draw (-35.5:6.2) edge[mathdbArrow={requires}] (-10.5:6.25) {};
    		
    		\node[mathdbCircle, below right=1cm and 4.5cm of matForm] (pub) {\large Publication};
    		\draw (pub.north) edge[mathdbArrow={* used in}] ($(pub.north) + (0, 3.5cm)$) {};
    		\draw (pub.north west) edge[mathdbArrow={\rotatebox{180}{* documented in}}] ($(pub.north west) + (-2.474cm, 2.474cm)$) {};
    		\draw ($(pub.west) + (-3.5cm, 0)$) edge[mathdbArrow={* invented in}] (pub.west) {};
    		\draw ($(pub.south west) + (-2.474cm, -2.474cm)$) edge[mathdbArrow={* surveyed in}] (pub.south west) {};
    		\draw ($(pub.south) + (0, -3.5cm)$) edge[mathdbArrow={* studied in}] (pub.south) {};
    		\node[text=red, below left=1cm and -1cm of bench] (mathalgodb) {\Large MathAlgoDB};
    		\node[text=blue, above left=1cm and -1cm of reseProb] (mathmoddb) {\Large MathModDB};
    		
    		\node[above left=5cm and -2.5cm of pub, align=right] {\large where * stands for:\\[2pt]\large all classes\\\\\large where ** may stand for:\\[2pt]\large formulation, assumption,\\[2pt]\large definition, initial condition,\\[2pt]\large final condition, boundary\\[2pt]\large condition, constraint condition\\\\\large where *** may stand for:\\[2pt]\large input, output, parameter,\\[2pt]\large constant, objective};
    	\end{tikzpicture}
     }
    \caption{Schematic display of the joint ontology for mathematical models and algorithms. Classes and important relations are depicted.     }
    \label{fig:JointOntologySchematic}
\end{figure}

In the preliminary work, MathModDB as the ontology for mathematical models and MathAlgoDB as the ontology for algorithms were rather separate developments. 
However, since they represent two sides of the same epistemic coin, they have been unified in this approach, see Fig.~\ref{fig:JointOntologySchematic}.

This is evident as the connections between the two parts are now made through several classes. First, a \textit{Computational Task} class had to be introduced, which is complementary to the \textit{Algorithmic Task} class and represents the semantic information content that a specific computational task is closely related to an algorithmic problem class. For example, the \textit{Computational Task} of calculating the time evolution of a dynamical system model is directly linked to the \textit{Algorithmic Task} of solving coupled ordinary differential equations.

Second, the \textit{Publication} class originally developed for MathAlgoDB is now also linked to MathModDB, since the reference information where a mathematical model (or any of its parts) was invented, analyzed, studied, surveyed or used in the literature is a crucial information for researchers.

Third, since the properties of mathematical formulations and the quantities contained in them are highly relevant for the selection of potential solution algorithms, a corresponding relation between \textit{Mathematical Formulation} and \textit{Algorithmic Task} was introduced.

\subsection{Quantities are central to semantic Knowledge Representation}
\label{subsec:Quantities}
As stated in the previous section, quantities play a decisive role in mathematical expressions as they give models their actual semantic meaning. Hence, next to the \textit{Quantity} class, the \textit{Quantity Kind} class was introduced. The rationale behind this distinction is to enhance clarity and precision in our semantic representations by categorically separating basic quantities from the specific quantities that occur in the use cases. For example, the electrical membrane potential of a cell, which occurs as a quantity in one of the bio-physiological use cases, is a specialization of the quantity \textit{Electrical Potential}. In electrostatics, the difference in this potential between two points is given by the basic quantity \textit{Voltage}, which is an individual of the \textit{Quantity Kind}  class. \textit{Voltage} then refers to the corresponding entries in the controlled vocabulary QUDT\footnote{\url{https://qudt.org/vocab/quantitykind/Voltage}}~\cite{Foster2013,Keil2019} and in Wikidata\footnote{\url{https://www.wikidata.org/wiki/Q25428}}. 

\subsection{Metadata Enrichment for mathematical Models}
\label{subsec:MDModels}

The existing ontologies were revised with regard to the integration of external information sources, in particular controlled vocabularies. This is intended to drive the enrichment of the KG with metadata in an unambiguous way. The information was added either as data properties or annotation properties - depending on whether the information can be used to make new findings in reasoning or it should only be available to users. Individuals from the \textit{Quantity} class can now be equipped with IDs from QUDT. Only if quantities are clearly presented, potential risks of incorrect units or misinterpretation, which have already led to issues elsewhere\footnote{\url{https://www.simscale.com/blog/nasa-mars-climate-orbiter-metric/}}, can be avoided.

In addition, individuals in the \textit{Research Field} class can now be assigned identifiers from the German Research Foundation (DFG)\footnote{\url{https://www.dfg.de/en/research-funding/proposal-funding-process/interdisciplinarity/subject-area-structure}}, Mathematics Subject Classification (MSC)\footnote{\url{https://msc2020.org}}~\cite{Lange2012} and Physics Subject Headings (PhySH)\footnote{\url{https://physh.org/}}~\cite{Smith2020} classification systems. Furthermore, it is now possible to link individuals with their equivalents in Wikidata. 

Subject-specific metadata can also be integrated, such as a natural language description as a comment and links to Wikipedia or Wikidata as annotations for individuals of any class. Moreover, individuals of the {\em Mathematical Formulation} class can hold the specific mathematical expression in \LaTeX~or MathML, and all symbols are broken down into quantities. For all these quantities, relevant metadata can be attached, for instance marking whether a quantity is a scalar, vector, matrix or a higher order tensor, and, for example, to indicate its matrix structure and more special properties. Individuals of the {\em Mathematical Formulation} and the {\em Mathematical Model} class can both hold relevant properties as metadata, such as the order of the model equations, but also whether they are linear, convex, homogeneous, just to name a few.

 \subsection{Metadata Enrichtment for mathematical Algorithms}
 \label{subsec:MDAlgorithms}
One major goal of MathAlgoDB is the proposition of a catalog of appropriate numerical algorithms for a concrete application problem. As stated, mathematical modeling boils down to a computational task containing a formulation with quantities. The formulation will be general (take e.g. ``solve ordinary differential equation''), but the modeling will provide additional properties for the formulation (e.g. stiff/nonstiff, scalar, first order, linear) or the quantities involved (e.g. the system matrix $A$ in the formulation of a linear ODE $y'=Ay$ is symmetric) and are now available through MathModDB.

In order to use that information, several changes were made to MathAlgoDB. Most of these are related to algorithm selection for specific application tasks. We need to augment the information on algorithms that is currently present with information about (the properties of) the objects it is working on or is expected to produce. For instance in the case of linear equations, for the mathematical formulation of ``solve $Ax=b$ for $x$'':
\begin{itemize}
  \item General algorithms like $LU$ decomposition will work for all square and invertible matrices.
  \item Conjugate Gradient will require $A$ to be symmetric and positive definite (s.p.d.).
  \item If $A$ is Toeplitz, solving with Trench's algorithm is recommended.
  \item Expectation Maximization/Maximum likelihood type algorithms are generally applicable for problems with a positivity constraint on $A$ and $x$, but particularly recommended provided $b$ is Poisson--distributed.
\end{itemize}
In the preliminary approach, this was implemented by using named tasks and subtasks, defining a general {\bf linear equation task} and a {\bf linear equations for s.p.d. matrices} subtask. As we showed, this is an approach that works fine for purely theoretical considerations and follows the structure of typical textbook derivations of algorithms.
\par
However, this implies that no automatic assignment of models to computational tasks or algorithms is possible. The new approach defines properties for the quantities involved in an algorithm, and thus can use the information provided by the model. We give the example of a tomography problem. Since the underlying model is linear, after discretization, the problem will always reduce to a simple {\bf linear equation task}. However, using general algorithms would be a mistake.

        \begin{itemize}
                \item Computational Task: In the formulation of a linear system $Ax=b$, solve for~$x$.
                \item Property 1: The formulation stems from (a discretization of) the Radon transform. 
                \item Property 2: The Radon transform has been sampled using parallel beam geometry.
                \item Property 3: $A$ is sparse.
                \item (Plain) Filtered backprojection is an algorithm that requires the first two properties. It is {\bf recommended} in this case.
                \item Algebraic Reconstruction/Kaczmarz' Algorithm is an algorithm that requires property one, but does not make use of the sampling structure. It is {\bf possible}, but not recommended.
                \item (Plain) conjugate gradient will require $A$ to be symmetric, which is not the case here. It is {\bf unusable} (at least directly).
        \end{itemize}
        For this case, filtered backprojection should be selected. MathAlgoDB also provides links to various implementations.

This simple example shows that it is hardly possible to predefine available properties. Each community or application case will have to be able to define their own property set, which precludes us from defining this within the ontology.
\par
In order to materialize this idea in the graph, we link \emph{Algorithmic Task}s to the \emph{Mathematical Formulation} class by object properties. The formulations then contain data properties which are interpreted based on the object property. We define
\begin{itemize}
\item  {\bf requires}: All properties listed are required for the algorithm.
\item  {\bf recommends}: If the properties are satisfied, the algorithm is preferred.
\item  {\bf precludes}: When this property is present, the algorithm cannot be used.
\end{itemize}

\label{ex:odes}
    In the case of initial value problems for ODEs, Runge-Kutta methods are the primary choice. Some of them can handle stiffness, some cannot, and of course they differ by convergence order, in turn requiring smoothness of solutions of the ODE. In terms of the ontology:
    \begin{itemize}
        \item Forward Euler precludes stiffness.
        \item Backward Euler is recommended when the ODE is stiff.
        \item RK4 requires smoothness of (at least) order 4, precludes stiffness and is recommended when the smoothness is of order 4.
    \end{itemize}

For algorithm selection, an appropriate SPARQL query yields all algorithms and implementations that solve the mathematical task, satisfy all required properties, do not conflict with precluded properties, possibly sorted by the recommended property. The last one is important -- remember that the most general algorithms will always show up here, but will usually be inappropriate due to running time or precision.

\section{Living Knowledge Graph of Models \& Algorithms}
\label{sec:KnowledgeGraph}

As of mid August 2024, more than 120 mathematical models and over 200 algorithms have been added as individuals, 
turning the conceptual ontology into a living KG. 
The models come from more than twenty various research fields. Among others, there are models from continuum mechanics, semiconductor physics, enzyme kinetics or biophysics. Moreover, mathematical models used in archaeology and Egyptology are also included. 
Up to now, the current data corpus consists of manually maintained, curated information only. This is primarily due to the high demands placed on the data quality of an algorithm and model database. High data quality is of great importance for third parties to be able to rely on the information and to use the contained information for their own research or for verification purposes. 

\subsection{Use Case: From falling Apples to moving Planets}
\label{sec:UseCaseApple}
As a illustrative example, we consider the famous story of Sir Isaac Newton being inspired to formulate his theory of gravitation by watching the fall of an apple from a tree in the year 1666. 
We have implemented this within the MathModDB KG by including a \textit{Research Field} named ``Pomology'' (science of fruits) containing ``Gravitational Effects on Fruit'' as a specific \textit{Research Problem}. 
Currently, there is a choice of two \textit{Mathematical Model}s addressing this problem, i.e., ``Free Fall Models'' without and with the effect of air drag, both of them assuming constant gravitation.
The former one, actually dating back to Newton, contains the simple ``Free Fall Equation'', $\dot{v}=g$, as a \textit{Mathematical Formulation}. That equation contains the (time-derivative of the) free fall velocity $v$ and the gravitational acceleration $g$.
Obviously, these two \textit{Quantities} are assigned to the \textit{Quantity Kinds} velocity and acceleration, respectively.
The free fall model with air drag may be described by the nonlinear equation~$\dot{v}=g-\frac{\rho C_{D}Av^2}{2m}$ which, in addition to the case without drag, contains the density of air~$\rho$, the drag coefficient~$C_D$, the cross section~$A$ and the mass of the apple~$m$.

Within the context of these ``Free Fall Models'', several different \textit{Computational Task}s can be formulated, e.g., how long does it take for an apple to reach the ground, or with which velocity will it hit the ground.
While the above equations are so simple that they can be solved analytically, in more realistic free fall models the underlying equations have to be solved numerically, which falls into the realm of the MathAlgoDB knowledge graph containing suitable numerical solvers for ordinary differential equations such as the Runge Kutta family of algorithms.

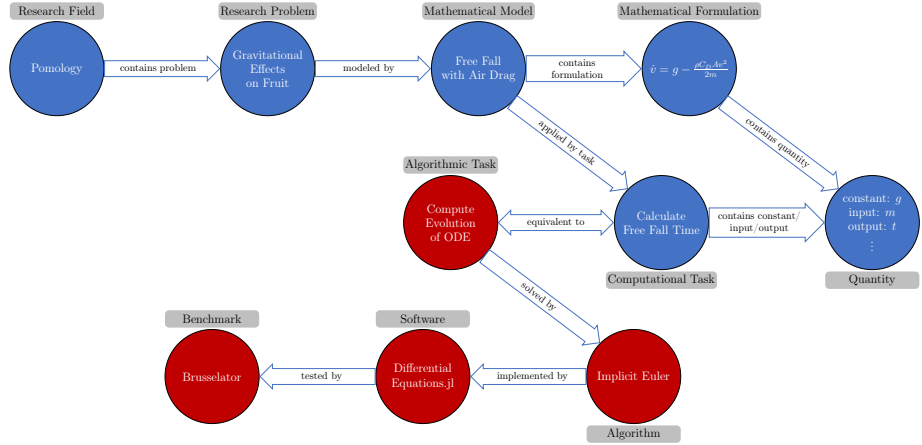
\begin{figure}[!ht]
    \centering
    \resizebox{\textwidth}{!}{
        \begin{tikzpicture}[node distance=3cm and 4cm]
    		\node[mathmoddbCircle] (reseField) {\large Pomology};
    		\node[mathmoddbCircle, right=of reseField] (reseProb) {\large Gravitational\\[2pt]\large Effects\\[2pt]\large on Fruit};
    		\node[mathmoddbCircle, right=of reseProb] (matModel) {\large Free Fall\\[2pt]\large with Air Drag};
    		\node[mathmoddbCircle, right=of matModel] (matForm) {\large $\dot{v}=g-\frac{\rho C_{D}Av^2}{2m}$};
    		\node[mathmoddbCircle, below right=of matModel] (compTask) {\large Calculate \\[2pt]\large Free Fall Time};
    		\node[mathalgodbCircle, left=of compTask] (algoTask) {\large Compute\\[2pt]\large Evolution\\[2pt]\large of ODE};
    		\node[mathmoddbCircle, right=of compTask] (quan) {\large constant: $g$\\[2pt]\large input: $m$\\[2pt]\large output: $t$\\[2pt]\large $\vdots$};
    		\node[mathalgodbCircle, below right=of algoTask] (algo) {\large Implicit Euler};
    		\node[mathalgodbCircle, left=of algo] (soft) {\large Differential\\[2pt]\large Equations.jl};
    		\node[mathalgodbCircle, left=of soft] (bench) {\large Brusselator};
    		
    		\node[labelBox, above=1.25pt of reseField] {\large Research Field};
    		\node[labelBox, above=1.25pt of reseProb] {\large Research Problem};
    		\node[labelBox, above=1.25pt of matModel] {\large Mathematical Model};
    		\node[labelBox, above=1.25pt of matForm] {\large Mathematical Formulation};
    		\node[labelBox, below=1.25pt of compTask] {\large Computational Task};
    		\node[labelBox, above=1.25pt of algoTask] {\large Algorithmic Task};
    		\node[labelBox, below=1.25pt of algo] {\large Algorithm};
    		\node[labelBox, above=1.25pt of soft] {\large Software};
    		\node[labelBox, above=1.25pt of bench] {\large Benchmark};
    		\node[labelBox, below=1.25pt of quan] {\large Quantity};
    
    		\draw (reseField.east) edge[mathdbArrow2={contains problem}] (reseProb.west) {};
    		\draw (reseProb.east) edge[mathdbArrow2={modeled by}] (matModel.west) {};
    		\draw (matModel.east) edge[mathdbArrow2={contains\hspace{1.5cm} formulation}] (matForm.west) {};
    		\draw (compTask.east) edge[mathdbArrow2={contains constant/ input/output}] (quan.west) {};
    		\draw (matModel.south east) edge[mathdbArrow2={applied by task}] (compTask.north west) {};
    		\draw (algoTask.east) edge[mathdbArrowEquiv={equivalent to}] (compTask.west) {};
    		\draw (algoTask.south east) edge[mathdbArrow2={solved by}] (algo.north west) {};
    		\draw (matForm.south east) edge[mathdbArrow2={contains quantity}] (quan.north west) {};
    		\draw (algo.west) edge[mathdbArrow2={\rotatebox{180}{implemented by}}] (soft.east) {};
    		\draw (soft.west) edge[mathdbArrow2={\rotatebox{180}{tested by}}] (bench.east) {};
    	\end{tikzpicture}
    }
    \caption{Schematic display of the sample use case~\ref{sec:UseCaseApple} free fall model in pomology in the joint ontology. For reasons of clarity, we will limit ourselves here to the essential entries and not show all connections, classes and quantities. }

    \label{fig:JointOntologyExample}
\end{figure}

In the case of the simple Free Fall Model in vacuum, virtually every scheme for solving ODEs will do. Following example~\ref{ex:odes} in~\ref{subsec:MDAlgorithms}, the simplest one (Forward Euler) will do its job if no order property is specified in the model.

For the air drag model~(see fig.~\ref{fig:JointOntologyExample}), we treat the equation as being stiff due to the nonlinearity. Since in the Forward Euler algorithm stiffness is a precluded property, it cannot be used in that case. However, Backward Euler is a possible algorithm for stiff ODEs, so it will be proposed here by MathAlgoDB.

Of course, for both examples, note that the model {\bf should} specify a smoothness class which would then trigger recommendation of a higher order Runge-Kutta scheme (such as RK4).
\par
The following SPARQL query can be used to find all algorithms together with respective \emph{Computational Task} and \emph{Algorithmic Problem} that can be applied when considering the \emph{Research Problem} of gravitational effects on fruits. Here, for simplicity, we only implement the ``precludes'' relation in order to exclude the explicit Runge-Kutta schemes for stiff equations:

{\scriptsize
\begin{lstlisting}
PREFIX madb: <https://mardi4nfdi.de/mathalgodb/0.1#>
PREFIX mmdb: <https://mardi4nfdi.de/mathmoddb#>
SELECT ?mod ?task ?prob ?form ?alg
WHERE {
  mmdb:GravitationalEffectsOnFruit mmdb:modeledBy ?mod .
  ?task mmdb:appliesModel ?mod .
  ?task mmdb:equivalentTo ?prob .
  ?form mmdb:containedAsFormulationIn ?mod .
  ?alg madb:solves ?prob .
  FILTER (  # implements the 'precludes' relation
    NOT EXISTS {
      ?alg madb:precludes ?precForm .
      FILTER (
        NOT EXISTS {
          ?precForm ?a ?b .
          FILTER (CONTAINS(STR(?a), STR(mmdb:)))
          FILTER (
            NOT EXISTS {
              ?form ?a ?b .
            })})})}
\end{lstlisting}
}

This query gives the expected output:
\begin{table}[ht]
    \centering
    \tiny
    \begin{tabular}{c|c|c|c|c}
        mod & task & prob & form & alg \\
        \hline
        FreeFallModelAirDrag & FreeFallDetermineVelocity & ComputeEvolutionODE & FreeFallEquationAirDrag & RKim11 \\
        FreeFallModelVacuum & FreeFallDetermineVelocity & ComputeEvolutionODE & FreeFallEquationVacuum & RKex11 \\
        FreeFallModelVacuum & FreeFallDetermineVelocity & ComputeEvolutionODE & FreeFallEquationVacuum & RKim11 \\
        FreeFallModelVacuum & FreeFallDetermineVelocity & ComputeEvolutionODE & FreeFallEquationVacuum & RK44kutta
    \end{tabular}
\end{table}

\subsection{Use Case: Like a Virus - Romanization of Northern Tunesia}
\label{sec:UseCaseImperialism}
From 146 BC to 350 AD, Roman culture spread across Northern Tunisia. Today, this can be traced by looking at archaeological evidence and historical records such as the administrative status of cities. In the \textit{Research Field} of Roman Archaeology, this phenomenon is studied under the \textit{Research Problem} of ``Romanization Spreading in Northern Tunisia''.

In~\cite{Kostre2022}, the spreading was modeled using a susceptible-infected (SI) model from the epidemiology domain on interregional networks with a time-dependent spreading rate. In this \textit{Mathematical Model} the state of each region $m$ at time $t$ is given by $s_m(t)$, the number of susceptible (non-Romanized) cities, and $i_m(t)$, the number of infected (Romanized) cities. The change in the number of susceptible and infected cities over time is described by the following \textit{Mathematical Formulation}s,
\begin{align*}
    \frac{d s_m(t)}{dt} &= -s_m(t) \alpha(t) \sum_{n=1}^{N_R} G_{m,n} i_n(t)\\
    i_m(t) &= P_m - s_m(t)
\end{align*}

with $\alpha(t)$, $G$, $P_m$ and $N_R$ denoting the \textit{Quantities} time-dependent spreading rate, interregional contact network, total number of cities in region $m$, and total number of regions, respectively.

Two \textit{Computational Task}s are associated with the model. These are: 1) the determination of approximate spreading curves $\phi(t)$ using arbitrary parameters for $G$ and $\alpha(t)$ in a first-order Runge-Kutta scheme and 2) the determination of optimal parameters $G$ and $\alpha(t)$ with respect to the real-world data using the Prescaled Metropolis-Adjusted Langevin Algorithm to minimize a loss function.

The initial value ODE problem in MathAlgoDB can be handled as in the Free Fall Model. 
Properties of the problem that would hint to use the Langevin Algorithm are high dimension, smoothness and availability of strong gradient information.

\subsection{Data Flows in the Knowledge Graph}

Templates were developed that facilitate the process for researchers wishing to add mathematical models to the MathModDB in order to make these models FAIR. They are provided as Markdown files and are designed to have a low entry barrier, allowing users with little to no experience to get started easily while guaranteeing the important details of the mathematical model are gathered. The filled-in templates then serve as a basis for a further semantically annotated standardized description of the model employing the MathModDB ontology.

MaRDMO~\cite{Reidelbach2023} is an extension of the Research Data Management Organiser
~\cite{Engelhardt2017}, the most widely used tool for data management planning in Germany~\cite{Enke2023}, facilitating the documentation of interdisciplinary workflows. It offers a standardized questionnaire and integrates with resources like the MaRDI Portal\footnote{\url{https://portal.mardi4nfdi.de/wiki/Portal}}, Wikidata, and other knowledge bases. Utilizing the MathModDB ontology, MaRDMO allows the documentation of mathematical models either independently or as part of larger workflows. It facilitates seamless integration with the  KG by connecting existing model entities and creating new ones when necessary~\cite{Reidelbach2024}. This capability positions MaRDMO as a user-friendly and interdisciplinary interface to explore and contribute to the KG.

The MathAlgoDB KG can be accessed via a web interface which can be used to retrieve information and browse the graph. Furthermore, the web interface has recently been extended to allow for the addition of new data into the KG. The joint ontology including the examples shown above is available at \url{https://mtsr2024.m1.mardi.ovh/}. 

The SPARQL query shown in~sec.~\ref{sec:UseCaseApple} can be sent to the corresponding SPARQL port at \url{https://sparql.mtsr2024.m1.mardi.ovh/mathalgodb/query} in order to obtain the results presented above.

\section{Conclusion \& Outlook}
\label{sec:ConclusionOutlook}

We have presented extensions of the original design for semantically representing mathematical models and algorithms and merged two previously distinct ontologies (sec.~\ref{sec:MergingModAlgo}), becoming a living knowledge graph (sec.~\ref{sec:KnowledgeGraph}). The extensions include renaming due to harmonization effort (sec.~\ref{subsec:PrevWork}), the introduction of a \textit{Computational Task} class (sec.~\ref{subsec:CompTask}), the incorporation of controlled vocabularies for the \textit{Research Field} class and for the base quantities within a new class \textit{Quantity Kind}~(sec.~\ref{subsec:Quantities}) as well as enabling and implementation of metadata enrichment for both models~(sec.~\ref{subsec:MDModels}) and algorithms~(sec.~\ref{subsec:MDAlgorithms}). With these, we are now able to semantically represent the essential parts of modeling and simulation as they occur in many mathematized branches of science.
More complex models than simple free fall models presented in sec.~\ref{sec:UseCaseApple} have been implemented, such as the dynamics of planets in our solar system.
This hints to a potential future work introducing a class encompassing theoretical methods or concepts such as the gravitational theory and classical mechanics.

We have already incorporated more than 250 models and algorithms but as of now, there are almost exclusively research assets for use cases from applied and numerical mathematics. In our future work we will investigate how the joint ontology can be adapted to other areas of mathematics, especially those reflected in MaRDI and the NFDI. 
Limitations of our approach include the handling of ``discretization''. Although it is an essential part of modeling and simulation, it is only implicitly represented in the current approach on the side of the models or on the side of the algorithms, depending on the specific use case. Another limitation at the moment is manual data ingest: although this guarantees high data quality, it is tedious and time-consuming. Strategies are to be developed to partially automate this process. Within the MaRDI project, the KG is to be incorporated into a MaRDI portal in which persistent IDs for models and algorithms can be assigned.

\subsection*{Acknowledgements}
The co-authors B.S., C.B., J.F., M.R., A.S., B.Sch. acknowledge funding by MaRDI, funded by the DFG (German Research Foundation), project number 460135501, NFDI 29/1 “MaRDI – Mathematische Forschungsdateninitiative”. The co-authors F.W. and H.K. acknowledge funding by the DFG under Germany's Excellence Strategy EXC 2044-390685587, Mathematics Münster: Dynamics–Geometry–Structure. The co-author D.G. acknowledges funding by the DFG under Germany's Excellence Strategy EXC 2075:  Data-Integrated Simulation Science (SimTech), project number 390740016.

 \bibliographystyle{splncs04}
  \bibliography{MTSR24_preprint}

\begin{thebibliography}{10}
\providecommand{\url}[1]{\texttt{#1}}
\providecommand{\urlprefix}{URL }
\providecommand{\doi}[1]{https://doi.org/#1}

\bibitem{Arndt2022_m4i}
Arndt, S., Farnbacher, B., Fuhrmans, M., Hachinger, S., Hickmann, J., Hoppe,
  N., Horsch, M.T., Iglezakis, D., Karmacharya, A., Lanza, G., Leimer, S.,
  Munke, J., Terzijska, D., Theissen-Lipp, J., Wiljes, C., Windeck, J.:
  Metadata4ing: An ontology for describing the generation of research data
  within a scientific activity. (2023). \doi{10.5281/zenodo.5957104}

\bibitem{Auer2018}
Auer, S., Kovtun, V., Prinz, M., Kasprzik, A., Stocker, M., Vidal, M.E.:
  {Towards a knowledge graph for science}. In: Proceedings of the 8th
  international conference on web intelligence, mining and semantics. pp.~1--6
  (2018), \url{http://doi.org/10.1145/3227609.3227689}

\bibitem{Baker2016}
Baker, M.: 1,500 scientists lift the lid on reproducibility. Nature
  \textbf{533}(7604),  452--454 (2016), \url{http://doi.org/10.1038/533452a}

\bibitem{Boege2022}
Boege, T., Fritze, R., G{\"o}rgen, C., Hanselman, J., Iglezakis, D., Kastner,
  L., Koprucki, T., Krause, T.H., Lehrenfeld, C., Polla, S., et~al.: {Data
  Management Planning in the German Mathematical Community}. European
  Mathematical Society Magazine (130),  40--47 (2023),
  \url{http://doi.org/10.4171/mag/152}

\bibitem{Chelliah2013}
Chelliah, V., Laibe, C., Le~Nov{\`e}re, N.: Biomodels database: a repository of
  mathematical models of biological processes. In Silico Systems Biology pp.
  189--199 (2013), \url{http://doi.org/10.1007/978-1-62703-450-0_10}

\bibitem{Conrad2024}
Conrad, T.O., Ferrer, E., Mietchen, D., Pusch, L., Stegm{\"u}ller, J.,
  Schubotz, M.: {Making Mathematical Research Data FAIR: Pathways to Improved
  Data Sharing}. Scientific Data  \textbf{11}(1), ~676 (2024),
  \url{http://doi.org/10.1038/s41597-024-03480-0}

\bibitem{Dutta2022}
Dutta, B., Patel, J.: Algorithm metadata vocabulary: A representational model
  and metadata vocabulary for describing and maintaining algorithms. Journal of
  Information Science  (2022), \url{http://doi.org/10.1177/01655515221116557}

\bibitem{Dyvak2022}
Dyvak, M., Melnyk, A., Rot, A., Hernes, M., Pukas, A.: Ontology of mathematical
  modeling based on interval data. Complexity  \textbf{2022},  1--19 (2022),
  \url{http://doi.org/10.1155/2022/8062969}

\bibitem{Elizarov2017}
Elizarov, A., Kirillovich, A., Lipachev, E., Nevzorova, O.: {Digital ecosystem
  OntoMath: Mathematical knowledge analytics and management}. In: Data
  Analytics and Management in Data Intensive Domains: XVIII International
  Conference, DAMDID/RCDL 2016, Ershovo, Moscow, Russia, October 11-14, 2016,
  Revised Selected Papers 18. pp. 33--46. Springer (2017),
  \url{http://doi.org/10.1007/978-3-319-57135-5_3}

\bibitem{Engelhardt2017}
Engelhardt, C., Enke, H., Klar, J., Ludwig, J., Neuroth, H.: Research data
  management organiser. In: Proceedings of the 14th International Conference on
  Digital Preservation. pp. 25--29 (2017)

\bibitem{Enke2023}
Enke, H., Hausen, D., Henzen, C., Jagusch, G., Krause, C., Sch\"onau, S.,
  et~al.: Data management planning: Concept for setting up a working group in
  the nfdi section common infrastructures. Zenodo  (2023),
  \url{http://doi.org/10.5281/zenodo.7540682}

\bibitem{Esteves2015}
Esteves, D., Moussallem, D., Neto, C., Soru, T., Usbeck, R., Ackermann, M.,
  Lehmann, J.: Mex vocabulary: A lightweight interchange format for machine
  learning experiments (2015), \url{http://doi.org/10.1145/2814864.2814883}

\bibitem{Falileeva2020}
Falileeva, M.V., Kirillovich, A., Shakirova, L.R., Nevzorova, O., Lipachev,
  E.K., Dyupina, A.: {OntoMathEdu Educational Mathematical Ontology:
  Prerequisites, Educational Levels and Educational Projections}. In: SSI. pp.
  346--351 (2020)

\bibitem{Foster2013}
Foster, M.P.: Quantities, units and computing. Computer Standards \& Interfaces
   \textbf{35}(5),  529--535 (2013),
  \url{http://doi.org/10.1016/j.csi.2013.02.001}

\bibitem{Hartl2021}
Hartl, N., W{\"o}ssner, E., Sure-Vetter, Y.: Nationale
  forschungsdateninfrastruktur (nfdi). Informatik Spektrum  \textbf{44}(5),
  370--373 (2021), \url{http://doi.org/10.1007/s00287-021-01392-6}

\bibitem{Inizan2021}
Inizan, O., Fromion, V., Goelzer, A., Sa{\"\i}s, F., Symeonidou, D.: {An
  Ontology to structure Biological Data: The Contribution of Mathematical
  Models}. In: Research Conference on Metadata and Semantics Research. pp.
  57--64. Springer (2021), \url{http://doi.org/10.1007/978-3-030-98876-0_5}

\bibitem{Keil2019}
Keil, J.M., Schindler, S.: Comparison and evaluation of ontologies for units of
  measurement. Semantic Web  \textbf{10}(1),  33--51 (2019),
  \url{http://doi.org/10.3233/SW-180310}

\bibitem{Kirillovich2021}
Kirillovich, A., Falileeva, M., Nevzorova, O., Lipachev, E., Dyupina, A.,
  Shakirova, L.: Prerequisite relationships of the ontomathedu educational
  mathematical ontology. In: Figueroa-Garc{\'i}a, J.C., D{\'i}az-Gutierrez, Y.,
  Gaona-Garc{\'i}a, E.E., Orjuela-Ca{\~{n}}{\'o}n, A.D. (eds.) Applied Computer
  Sciences in Engineering. pp. 517--524. Springer International Publishing
  (2021), \url{http://doi.org/10.1007/978-3-030-86702-7_44}

\bibitem{Kohlhase:OMDoc}
Kohlhase, M.: {OMDoc} -- An open markup format for mathematical documents
  [Version 1.2], LNAI, vol.~4180. Springer Verlag (2006),
  \url{http://doi.org/10.1007/11826095}

\bibitem{Koprucki2018}
Koprucki, T., Kohlhase, M., Tabelow, K., M{\"u}ller, D., Rabe, F.: {Model
  Pathway Diagrams for the Representation of Mathematical Models}. Optical and
  Quantum Electronics  \textbf{50}, ~1--9 (2018),
  \url{http://doi.org/10.1007/s11082-018-1321-7}

\bibitem{Kostre2022}
Kostré, M., Sunkara, V., Schütte, C., Conrad, N.D.: Understanding the
  romanization spreading on historical interregional networks in northern
  tunisia. Applied Network Science  \textbf{7}(53) (2022),
  \url{http://doi.org/10.1007/s41109-022-00492-w}

\bibitem{Lange2012}
Lange, C., Ion, P., Dimou, A., Bratsas, C., Sperber, W., Kohlhase, M.,
  Antoniou, I.: Bringing mathematics to the web of data: The case of the
  mathematics subject classification. In: Simperl, E., Cimiano, P., Polleres,
  A., Corcho, O., Presutti, V. (eds.) The Semantic Web: Research and
  Applications. pp. 763--777. Springer Berlin Heidelberg (2012),
  \url{http://doi.org/10.1007/978-3-642-30284-8_58}

\bibitem{Nguyen2020}
Nguyen, A., Weller, T., Färber, M., Sure-Vetter, Y.: Making {Neural Networks
  FAIR} (2020), \url{https://arxiv.org/abs/1907.11569}

\bibitem{Reidelbach2023}
Reidelbach, M., Ferrer, E., Weber, M.: {MaRDMO Plugin - Document and Retrieve
  Workflows Using the MaRDI Portal}. In: {Proceedings of the 1st Conference on
  Research Data Infrastructure (CoRDI) - Connecting Communities } (2023).
  \doi{10.52825/cordi.v1i.254}

\bibitem{Reidelbach2024}
Reidelbach, M., Schembera, B., Weber, M.: Towards a fair documentation of
  workflows and models in applied mathematics. In: Buzzard, K., Dickenstein,
  A., Eick, B., Leykin, A., Ren, Y. (eds.) Mathematical Software -- ICMS 2024.
  pp. 254--262. Springer Nature Switzerland (2024),
  \url{http://doi.org/10.1007/978-3-031-64529-7_27}

\bibitem{Riedel2022}
Riedel, C., Geßner, H., Seegebrecht, A., Ayon, S.I., Chowdhury, S.H., Engbert,
  R., Lucke, U.: Including data management in research culture increases the
  reproducibility of scientific results. INFORMATIK 2022 (2022),
  \url{http://doi.org/10.18420/inf2022_114}

\bibitem{Sack2023}
Sack, H., Schrade, T., Bruns, O., Posthumus, E., Tietz, T., Norouzi, E.,
  Waitelonis, J., Söhn, L., Fliegl, H., Tolksdorf, J., Jalle-Steller, J.,
  Guzmán, A.A., Fathalla, S., Ihsan, A.Z., Hofmann, V., Sandfeld, S., Fritzen,
  F., Laadhar, A., Schimmler, S., Mutschke, P.: {Knowledge Graph Based RDM
  Solutions}. In: {Proceedings of the 1st Conference on Research Data
  Infrastructure (CoRDI) - Connecting Communities } (2023),
  \url{http://doi.org/10.52825/cordi.v1i.371}

\bibitem{Schembera2020_DD}
Schembera, B., Dur{\'a}n, J.M.: Dark data as the new challenge for big data
  science and the introduction of the scientific data officer. Philosophy \&
  Technology  \textbf{33},  93--115 (2020),
  \url{http://doi.org/10.1007/s13347-019-00346-x}

\bibitem{schembera2024ontologies}
Schembera, B., W{\"u}bbeling, F., Kleikamp, H., Biedinger, C., Fiedler, J.,
  Reidelbach, M., Shehu, A., Schmidt, B., Koprucki, T., Iglezakis, D.,
  G{\"o}ddeke, D.: Ontologies for models and algorithms in applied
  mathematics and related disciplines. In: Garoufallou, E., Sartori, F. (eds.)
  Communications in Computer and Information Science. pp. 161--168. Springer
  Nature Switzerland, Cham (2024),
  \url{http://doi.org/10.1007/978-3-031-65990-4_14}

\bibitem{Schembera2023_CoRDI}
Schembera, B., Wübbeling, F., Koprucki, T., Biedinger, C., Reidelbach, M.,
  Schmidt, B., Göddeke, D., Fiedler, J.: {Building Ontologies and Knowledge
  Graphs for Mathematics and its Applications}. In: {Proceedings of the 1st
  Conference on Research Data Infrastructure (CoRDI) - Connecting Communities }
  (2023), \url{http://doi.org/10.52825/cordi.v1i.255}

\bibitem{Smith2020}
Smith, A.: {Physics Subject Headings (PhySH)}. KO KNOWLEDGE ORGANIZATION
  \textbf{47}(3),  257--266 (2020),
  \url{http://doi.org/10.5771/0943-7444-2020-3-257}

\bibitem{Snytnikov2020}
Snytnikov, A., Glinskiy, B., Zagorulko, G., Zagorulko, Y.: {Ontological
  Approach to Formalization of Knowledge in Computational PlasmaPphysics}.
  Journal of Physics: Conference Series  \textbf{1640},  012013 (2020),
  \url{http://doi.org/10.1088/1742-6596/1640/1/012013}

\bibitem{Suresh2010}
Suresh, P., Hsu, S.H., Akkisetty, P., Reklaitis, G.V., Venkatasubramanian, V.:
  {OntoMODEL: ontological mathematical modeling knowledge management in
  pharmaceutical product development, 1: conceptual framework}. Industrial \&
  Engineering Chemistry Research  \textbf{49}(17),  7758--7767 (2010),
  \url{http://doi.org/10.1021/ie100246w}

\bibitem{Suresh2008}
Suresh, P., Joglekar, G., Hsu, S., Akkisetty, P., Hailemariam, L., Jain, A.,
  Reklaitis, G., Venkatasubramanian, V.: {Onto MODEL: Ontological mathematical
  modeling knowledge management}. In: Computer Aided Chemical Engineering,
  vol.~25, pp. 985--990. Elsevier (2008),
  \url{http://doi.org/10.1016/S1570-7946(08)80170-8}

\bibitem{MaRDI2022}
{The MaRDI consortium}: {MaRDI: Mathematical Research Data Initiative Proposal}
  (2022). \doi{10.5281/zenodo.6552436}

\bibitem{Wilkinson2016}
Wilkinson, M.D., Dumontier, M., Aalbersberg, I.J., Appleton, G., Axton, M.,
  Baak, A., Blomberg, N., Boiten, J.W., da~Silva~Santos, L.B., Bourne, P.E.,
  et~al.: {The FAIR Guiding Principles for scientific data management and
  stewardship}. Scientific data  \textbf{3}(1), ~1--9 (2016),
  \url{http://doi.org/10.1038/sdata.2016.18}

\bibitem{Zang2022}
Zang, Z., Ma, T.: {Research and Application of Mathematical Knowledge Graph
  Based on Ontology Learning}. In: Liu, Q., Liu, X., Cheng, J., Shen, T., Tian,
  Y. (eds.) Proceedings of the 12th International Conference on Computer
  Engineering and Networks. pp. 1387--1394. Springer Nature Singapore (2022),
  \url{http://doi.org/10.1007/978-981-19-6901-0_147}

\bibitem{Zwaneveld2000}
Zwaneveld, B.: Structuring mathematical knowledge and skills by means of
  knowledge graphs. International Journal of Mathematical Education in Science
  and Technology  \textbf{31}(3),  393--414 (2000),
  \url{http://doi.org/10.1080/002073900287165}

\end{thebibliography}
\end{document}